%% file: neurips_2025.tex
\documentclass{article}

    \usepackage[preprint]{neurips_2025}

\usepackage[utf8]{inputenc} % allow utf-8 input
\usepackage[T1]{fontenc}    % use 8-bit T1 fonts
\usepackage{hyperref}       % hyperlinks
\usepackage{url}            % simple URL typesetting
\usepackage{booktabs}       % professional-quality tables
\usepackage{amsfonts}       % blackboard math symbols
\usepackage{nicefrac}       % compact symbols for 1/2, etc.
\usepackage{microtype}      % microtypography
\usepackage{xcolor}         % colors
\usepackage{amsmath}
\usepackage{graphicx}
\usepackage{cleveref}
\definecolor{RL}{RGB}{255, 0, 0}
\definecolor{CGS}{RGB}{0, 0, 255}

\title{Reinforcement Learning Closures for Underresolved Partial Differential Equations using Synthetic  Data}

\author{
  \textbf{Lothar Heimbach}\\
  ETH Zurich/ Harvard University\\
  \texttt{lotharheimbach5@gmail.com}
  \And
  \textbf{Sebastian Kaltenbach}\\
  Harvard University\\
  \texttt{skaltenbach@seas.harvard.edu}
  \And
 \textbf{Petr Karnakov}\\
  Harvard University\\
  \texttt{pkarnakov@seas.harvard.edu}
  \And
  \textbf{Francis J. Alexander}\\
  Argonne National Laboratory\\
  \texttt{fja@anl.gov}
  \And
  \textbf{Petros Koumoutsakos}\\ 
  Harvard University\\
  \texttt{petros@seas.harvard.edu}
}

\begin{document}

\maketitle

\begin{abstract}
Partial Differential Equations (PDEs) describe phenomena ranging from turbulence and epidemics to quantum mechanics and financial markets.  Despite recent advances in computational science, solving such PDEs for real-world applications remains prohibitively expensive because of the necessity of resolving a broad range of spatiotemporal scales. In turn, practitioners often rely on coarse-grained approximations of the original PDEs, trading off accuracy for reduced computational resources. To mitigate the loss of detail inherent in such approximations, closure models are employed to represent unresolved spatiotemporal interactions. We present a framework for developing closure models for PDEs using synthetic data acquired through the method of manufactured solutions. These data are used in conjunction with reinforcement learning to provide closures for coarse-grained PDEs.  We illustrate the efficacy of our method using the one-dimensional and two-dimensional Burgers' equations and the two-dimensional advection equation. Moreover,  we demonstrate that closure models trained for inhomogeneous PDEs can be effectively generalized to homogeneous PDEs. The results demonstrate the potential for developing accurate and computationally efficient closure models for systems with scarce data. 
\end{abstract}

\section{Introduction}
% Reference closure-rl but how it was trained on FGS. taking up ~70% of training time
% Reference novati automated turbulence paper and how they tried to use germano identity as a FGS independent reward, but did not yield superior results to to static-/dynamic-Smagorinksy model. Still forced to really on FGS (I think..)
%Reference paper Sebastian found using the MMS

The accurate solution of partial differential equations (PDEs) is a cornerstone of computational science, playing a critical role in applications such as weather forecasting \citep{ehrendorfer1997predicting,dunlea2012national}, turbulence modeling \citep{wilcox1988multiscale,spalart2000strategies}, and traffic planning \citep{lighthill1955kinematic}. Precise PDE solutions enable the understanding of complex physical phenomena as well as optimization of engineering processes and efficiency. However, achieving high accuracy in PDE simulations remains a significant challenge, particularly for real-world engineering applications, where resolving all relevant scales demands exceedingly fine grids and small time steps, leading to prohibitive computational costs \citep{rossinelli201311,palmer2015modelling}.

For tubulence modeling, the cost of direct numerical simulations scales steeply with the Reynolds number, making it infeasible for high Reynolds number flows encountered in engineering and atmospheric sciences. To address this challenge, reduced-order modeling techniques, such as Large Eddy Simulation (LES) \citep{deardorff1970numerical}, have emerged as practical alternatives by resolving only the largest turbulent structures while modeling the effects of unresolved subgrid scales. However, the reliability of LES is highly dependent on the choice of the subgrid-scale (SGS) model, which remains an active research area. Although traditional SGS models are based on empirically derived approximations \citep{smagorinsky1963general}, \citep{maulik2018explicit}, recent advances in machine learning have paved the way for data-driven SGS models \citep{brunton2020machine}, offering the potential for more accurate and adaptable turbulence modeling approaches.

In recent years, a multitude of machine learning-based methods have been proposed to tackle the challenges of solving PDEs both accurately and efficiently. This includes approaches based on learning a suitable latent space \citep{kingma2013auto,chen2018neural,champion2019data,yildiz2019ode2vae,kaltenbach2021physics,vlachas2022multiscale,kivcic2023adaptive,gao2024generative,zang2025dgno} as well as neural operator frameworks that directly approximate the solution operator of a PDE \citep{lu2021learning,kovachki2023neural,li2020fourier}. Moreover, approaches that complement existing coarse-grained simulations or physical models by learning closure models have gained popularity \citep{novati2021,bae2022scientific,vonbassewitz2024closure}. However, to train these methods, solutions of the underlying fully resolved PDE have to be computed, leading to long training times and a high demand for data, which is a major drawback.

The amount of training data can be reduced by incorporating additional knowledge into the loss function or machine learning architecture. Popular approaches to enforce physical constraints include a physics-informed loss term \citep{raissi2019physics,karniadakis2021physics,wang2021learning,cai2021physics,kaltenbach2023semi,karnakov2024solving} or virtual observables \citep{kaltenbach2020incorporating,rixner2021probabilistic,scholz2025weak} to incorporate the governing physics of the system directly into the learning process and eliminate the need for large amounts of costly training data. Alternatively, physics-inspired neural network architectures such as Hamiltonian Neural Networks \citep{greydanus2019hamiltonian,toth2019hamiltonian} or Lagrangian Neural Networks \citep{lutter2019deep,cranmer2020lagrangian} can be employed.

In this work, we propose a novel approach to mitigate the need for large amounts of training data. Using synthetic training data generated using the method of manufactured solutions (MMS) \citep{roache2002code}, we eliminate the need for costly evaluations of the original PDE system during training while retaining all the advantages of data-driven learning. Within this paper, we employ a RL based approach to learn a closure model for coarse-grained PDEs that generalizes to unseen test cases.

Recent research by \cite{hasani2024generatingsyntheticdataneural} employed synthetic training data to train a Fourier Neural Operator (FNO). In contrast, this work uses synthetic data to train a general data-driven closure model. Furthermore, we demonstrate that this closure model can be applied to homogeneous PDEs, allowing extrapolative test cases not explicitly included within the synthetic MMS data. We illustrate the approach on the 1D Burgers' equation, the 2D Burgers' equation, and the 2D advection equation, underscoring the closure models’ capacity to enhance the accuracy of coarse-grained PDE solutions without the substantial computational expense typically associated with model training.
%The remainder of this paper is structured as follows. In Section \ref{Sec:method}, we introduce the methodology, encompassing the framework for generating synthetic training data based on the MMS and the reinforcement learning framework employed for closure modeling. Subsequently, we present three numerical illustrations in Section \ref{sec:numerical_illustration}. Finally, Section \ref{sec:conclusions} presents the conclusions of this study.

\section{Methodology}
\label{Sec:method}
We present training of a RL-based closure modeling framework for coarse-grained PDEs based on synthetic training data generated by the method of manufactured solutions (MMS) described in Section \ref{sec:mms}. Subsequently, we introduce the coarse-grained simulation and the closure modeling approach in Section \ref{sec:cgs}. Finally, in Section \ref{sec:RL}, we present the reinforcement learning (RL) framework together with the relevant training methodologies and implementation details.
\subsection{Method of Manufactured Solutions}
\label{sec:mms}
The MMS technique, as described in \cite{roache2002code}, was initially developed as a method to verify the precision and stability of the numerical methods used to solve partial differential equations (PDEs). In MMS, an analytical solution, known as the 'manufactured solution', is selected and substituted into the PDE, resulting in the introduction of a novel source or forcing term. This approach enables programmers to create a known solution and assess its performance without directly solving the PDE. MMS finds particular utility in testing intricate numerical schemes, as it provides a controlled environment with precise solutions for benchmarking and error analysis. Furthermore, MMS facilitates the analytical expression of the solution in a PDE for which an analytical solution may not otherwise be available.\\
In this paper, we will not employ MMS to evaluate an implementation of a numerical method, but rather to generate solutions for a known PDE. It is important to note that each generated solution will be associated with its own distinct forcing term. To illustrate the application of MMS, we will present a demonstration on a simplified example that involves the 1D Burgers' equation as the governing PDE.

The Burgers' equation with solution $\psi(x,t)$ is defined as:
\begin{equation}
    \psi_t +\psi\psi_x - \nu \psi_{xx} = \mathcal{L}(\psi(x,t))
\label{eq:general_PDE_form}
\end{equation}
Here, $\mathcal{L}(\psi(x,t))$ is a general forcing term that would be set to zero in case of the homogeneous Burgers equation and we use subindices $t$ and $x$ to indicate the respective derivatives of the solution function.\\
Next we specify a proposed solution which is differentiable up to at least the same order as appears in the derivatives appearing in the PDE. In the case of the 1D Burger's equation, the proposed solution has to be first-order differentiable w.r.t. $t$ and second-order differentiable w.r.t $x$. For simplicity, we choose the proposed solution
\begin{equation}
    \psi_\text{MMS}(x,t) = t \cdot\cos(x).
\label{eq:toy_proposed_solution}
\end{equation}
With the solution being predetermined we need to calculate the forcing term $\mathcal{L}(\psi(x,t)_\text{MMS})$ such that equation \ref{eq:general_PDE_form} is true. Given the PDE and proposed solution, we substitute the proposed solution into the PDE and simplify the equation.

\begin{align}
    \mathcal{L}(\psi_\text{MMS}(x,t)) &= (\psi_\text{MMS})_t + \psi_\text{MMS}(\psi_\text{MMS})_x - \nu (\psi_\text{MMS})_{xx}  \notag \\
     &= (t\cdot\cos(x))_t + t\cdot\cos(x) (t\cdot\cos(x))_x - \nu (t\cdot\cos(x))_{xx}  \\
     &=  \cos(x) - t\cdot\cos(x) \cdot t \cdot \sin(x) + \nu t\cdot\cos(x)  \notag
\label{eq:general_forcing_term}
\end{align}
Subsequently, we have obtained the forcing term corresponding to the manufactured solution and with the initial condition set to the proposed solution at $t=0$, we obtain the following initial value problem:
\begin{equation} 
    \begin{cases} 
    \psi_t +\psi\psi_x - \nu \psi_{xx} = \mathcal{L}(\psi_\text{MMS}(x,t)) \\
    \psi(x,0) = \psi_\text{MMS}(x,0) \\
    x\in[0,L),\; t\in[0,T]
    \end{cases}
\label{eq:toy_IBVP}
\end{equation}
The solution to the initial value problem described in equation \ref{eq:toy_IBVP} is $\psi_\text{MMS}(x,t)$. Any error arising during the numerical computation comes from the approximation error of the numerical methods employed.

We note that as we are using the MMS to obtain synthetic training data, we are free to choose any $\psi_\text{MMS}(x,t)$. To obtain a diverse set of synthetic data, we opted to choose parametrized functions for $\psi_\text{MMS}(x,t)$ as specified in section \ref{sec:numerical_illustration}.

\subsection{Coarse and Fine Grid Simulation}
\label{sec:cgs}
Within this work, we try to solve PDEs and obtain a discretized solution both with minimal error and with minimal computation time. Although fully resolving all scales of a spatio-temporal PDE of interest by using a fine-grid simulation (FGS) would lead to minimal error, this also implies a high computational cost. We are thus employing a coarse-grid simulation (CGS) to reduce computation time and are learning a closure scheme to obtain sufficient accuracy.\\
This CGS entails a coarse spatial discretization $ \widetilde{\Delta x}= d \Delta x$, $ \widetilde{\Delta y}= d \Delta y$, $ \widetilde{\Delta z}= d \Delta z$ as well as a coarse temporal discretization $ \widetilde{\Delta t} = d_t \Delta t$. Here, $d$ is the spatial scaling factor and $d_t$ the temporal scaling factor, and $\Delta t$ and $\Delta x, \Delta y, \Delta z$ are discretizations that resolve all spatio-temporal scales of interest for the spatial dimensions $x,y,z$.\\
In a time step $n$ the discretized solution $\tilde{\boldsymbol{\psi}} ^{n}$ of the CGS can be computed using an update rule based on the solution in the previous time step and the parameters of the PDE $\tilde{C}$. 
\begin{equation}
\label{eq:cg_dyn}
\tilde{\boldsymbol{\psi}}^{{n}+1} =\mathcal F(\tilde{\boldsymbol{\psi}}^{n}, \tilde C^{n}).
\end{equation}
This discretized solution of CGS is a subsampled version of the FGS solution $\boldsymbol{\psi}^{{n}}$  and the subsampling operator $\mathcal S:  \Psi \rightarrow \tilde \Psi $ connects the two.
We note that we generate the FGS solution with the MMS and are never required to actually simulate the FGS during training of the closure model.

\subsection{RL Framework}
\label{sec:RL}
\begin{figure}[ht]
\vskip 0.2in
\begin{center}
\includegraphics[width=1.0\linewidth]{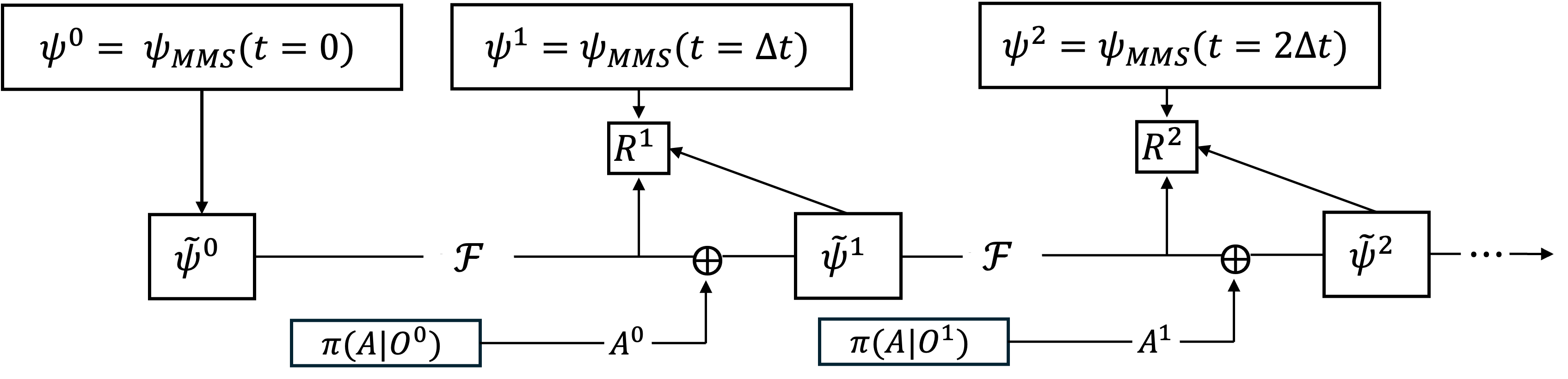}
\caption{Illustration of the RL framework with the agents embedded in the CGS and their action $A$ computed by a policy $\pi$. The CGS solution is computed by using the update operator $\mathcal{F}$ and afterwards modified by the agents in order to bring the solution closer to the FGS. The reward $R$ measures how much the action taken improves the CGS.}
\label{fig:environment}
\end{center}
\vskip -0.2in
\end{figure}

In this work, we employ the recently introduced Closure-RL framework \citep{vonbassewitz2024closure}. The framework and the environment were modified to utilize the additional information introduced by the MMS in order to improve its performance. The RL framework is summarized in \ref{fig:environment}. \\
We define the state at step $n$ of the RL environment as the tuple $\boldsymbol{S}^n := (\boldsymbol{\psi}^n, \tilde{ \boldsymbol{\psi}^n},\tilde{ \boldsymbol{\psi}^{n-1}},\tilde{C}^n)$. This state is only partially observable as the policy is acting only in the CGS. The observation \(O^n = \left( \frac{\mathcal{L}(\tilde{\psi}^n)}{\max|\mathcal{L}(\tilde{\psi}^n)|}, \Tilde{\psi}^{n-1}, \Tilde{\psi}^{n}, \tilde{C}^n \right)\) is defined as the coarse solutions of the last two time steps as well as the normalized forcing term of the PDE (see equation \ref{eq:general_PDE_form}) and the PDE parameters.\\
the goal is to train a policy $\pi$ that complements the dynamics of the CGS in order to be as close as possible to the dynamics of  the FGS. To achieve this goal, the action ${\boldsymbol{ A}^n}$ at step $n$ of the environment is a collection of forcing terms for each discretization point of the CGS. In case the policy is later used during predictions, the update function in equation \ref{eq:cg_dyn} changes to 
\begin{equation}
   \tilde{\boldsymbol{\psi}}^{n+1}=\mathcal F(\tilde{\boldsymbol{ \psi}}^n , \tilde C^n)+ \boldsymbol{ A}^n. 
\end{equation}
To encourage learning a policy that represents the non-resolved spatio-temporal scales, the reward is based on the difference between the CGS and FGS at time step $n$. 
\begin{equation}
    R_i^n = [\mathcal{S}(\psi_i^n)-(\Tilde{\psi}_i^n)_{\text{Pre-Action}}]^2 - [\mathcal{S}(\psi_i^n)-\Tilde{\psi}_i^n]^2
\label{eq:reward}
\end{equation}

If the action \( \boldsymbol{A}^n \) moves the discretized solution function of the CGS, \( \tilde{\boldsymbol{\psi}^n} \), closer to the subsampled discretized solution function of the FGS, \( \mathcal{S}(\boldsymbol{\psi}^n) \), the reward is positive; otherwise, it is negative. \\
For parameterization as well as for the training of the policy, we use the architecture and optimization algorithm suggested in \cite{vonbassewitz2024closure}. In more detail, the RL policy is trained using an adapted version of proximal policy optimization (PPO) \citep{schulman2017proximal} implemented in the Tianshou library \citep{weng2022tianshou} .

\section{Numerical Illustrations}
\label{sec:numerical_illustration}

We present three examples of how the MMS can generate synthetic training data for RL to learn a general closure term for accurate predictions on both homogeneous and inhomogeneous PDEs.

%To iterate the numerical solution through time we define the time-stepping operator in equation \ref{eq:time_step_operator}.

\subsection{1D Burgers' Equation}
As a first example, we consider the 1D Burgers' equation on a domain \(\Omega^{\text{1D}} = [0,1]\) with periodic boundary conditions. 
\begin{equation}
\psi_t +\psi\psi_x - \nu \psi_{xx} = 0
\end{equation}
For the CGS, we use a time-stepping operator that employs the upwind scheme for first-order spatial derivatives, the central difference scheme for second-order spatial derivatives, and explicit Euler for the temporal derivative \citep{quarteroni2008numerical}.

To generate synthetic training data via the MMS, we employ Equation \ref{eq:mms_1D_Burgers} as the proposed solution to the 1D Burgers' equation:
\begin{equation}
    \label{eq:mms_1D_Burgers}
    \psi_\text{MMS}(x,t) = A\sin(a x - \delta t)e^{-ct} + B\cos(b x - \delta t)e^{-ct},
\end{equation}
where \( A, B \sim \text{Uniform}(0, 1) \); \( a, b \sim \{2\pi, 4\pi\} \); \( c \sim \text{Uniform}(0.1, 2) \), and \( \delta \sim \text{Uniform}(-2, 2) \). 

The proposed solution is periodic in $[0,1]$ for all $t$, ensuring agreement between the periodic boundary conditions and the solution at each time step. Moreover, it mimics the time-dependent evolution of the Burgers' equation. The exponential function acts as an artificial diffusion term, reducing the magnitude of the solution over time. A lower forcing term during training simplifies the generalization to the homogeneous case required by the learned closure model. The $-\delta t$ term inside the trigonometric functions acts as an artificial advection, translating the solution over time. We note that variations of this parameterized solution have been tested and did not significantly affect the solution.  

Details on the architecture choices for the RL framework and the hyperparameter can be found in the Appendix \ref{sec:numerical_parameters} and \ref{app:1D}. Moreover, visualization of training metrics are displayed in Appendix \ref{sec:tr1D}.

\input{1dBurger}

\subsection{2D Burgers' Equation}
As a second example, we consider the 2D Burgers' equation on a domain \(\Omega^{\text{2D}} = [0,1]\times[0,1]\) with periodic boundary conditions. 
\begin{equation}
\frac{\partial \boldsymbol{\psi}}{\partial t} + (\boldsymbol{\psi} \cdot \nabla) \boldsymbol{\psi} - \nu \nabla^2 \boldsymbol{\psi} = 0
\end{equation}
For the CGS, we again use a time-stepping operator that employs the upwind scheme for first-order spatial derivatives, the central difference scheme for second-order spatial derivatives, and explicit Euler for the temporal derivative \citep{quarteroni2008numerical}.

For the 2D Burgers' equation we define the MMS solution as \(\boldsymbol{\psi}_{\text{MMS}}=(u_\text{MMS},v_\text{MMS})\). The proposed solution is shown in equation \ref{eq:mms_2D_Burgers}:
\begin{equation}
\label{eq:mms_2D_Burgers}
\begin{aligned}
u_\text{MMS}(x, y, t) &= A_u \sin\left(2\pi x\right) \sin\left(2\pi y\right) \exp\left(-z_{\text{diff}, u} t\right) + B_u \cos\left(2\pi x\right) \sin\left(2\pi y\right) \exp\left(-z_{\text{diff}, u} t\right) \\
&\quad + C_u \sin\left(2\pi x\right) \cos\left(2\pi y\right) \exp\left(-z_{\text{diff}, u} t\right)
 + D_u \cos\left(2\pi x\right) \cos\left(2\pi y\right) \exp\left(-z_{\text{diff}, u} t\right), \\
v_\text{MMS}(x, y, t) &= A_v \sin\left(2\pi x\right) \sin\left(2\pi y\right) \exp\left(-z_{\text{diff}, v} t\right) + B_v \cos\left(2\pi x\right) \sin\left(2\pi y\right) \exp\left(-z_{\text{diff}, v} t\right) \\
&\quad + C_v \sin\left(2\pi x\right) \cos\left(2\pi y\right) \exp\left(-z_{\text{diff}, v} t\right) + D_v \cos\left(2\pi x\right) \cos\left(2\pi y\right) \exp\left(-z_{\text{diff}, v} t\right),
\end{aligned}
\end{equation}

where \( A_u, B_u, C_u, D_u \sim \text{Uniform}(0, 1) \); \( A_v, B_v, C_v, D_v \sim \text{Uniform}(0, 1) \); and \( z_{\text{diff}, u}, z_{\text{diff}, v} \sim \text{Uniform}(0.1, 3) \). Similarly to the 1D Burgers' equation before, the solution for 2D Burgers' equation enforces periodicity across $x\in[0,1]$ and also $y\in[0,1]$. There are now dedicated terms for each of the two spatial dimension as the proposed solution needs to be differentiable with regards to both spatial dimensions. Additionally, we now employ two artificial diffusion coefficients, one for each dimension. While we again have chosen trigonometric functions as the main building blocks for our parameterized proposed solution, we note that the solution to the homogenous 2D Burgers' equation is not expected to consist of trigonometric functions. 

Details on the architecture choices for the RL framework and the hyperparameter can be found in the Appendix \ref{sec:numerical_parameters} and \ref{app:2D}. Moreover, visualization of training metrics are displayed in Appendix \ref{sec:2D_Burgers_Training_Metrics}.

\input{2dBurger}

\subsection{2D Advection Equation}
As a third example, we consider the 2D Advection equation on a domain \(\Omega^{\text{2D}} = [0,1]\times[0,1]\) with periodic boundary conditions. 
\begin{equation}
\frac{\partial \psi}{\partial t} + u(x,y)\frac{\partial \psi}{\partial x} + v(x,y)\frac{\partial \psi}{\partial y} = 0,
\end{equation}
The velocity fields \(C=(u,v)\) are given by:
\begin{equation}
\label{eq:velocity_fields}
\begin{aligned}
u(x,y) &= x + \alpha \\
v(x,y) &= y + \beta
\end{aligned}
\end{equation}
where
\(\alpha, \beta \sim \text{Uniform}(-1, 1).\)
The CGS employs the same numerical schemes as for the other two examples.

The proposed solution of the 2D advection equation is given in equation \ref{eq:mms_2D_Advection}.
\begin{equation}
\label{eq:mms_2D_Advection}
\begin{aligned}
\psi_\text{MMS}(x, y, t) &= A \sin(a x ) \sin(b y )  + B \cos(c x) \sin(d y ) \\
&\quad + C \sin(e x ) \cos(f y )  + D \cos(g x ) \cos(h y )
\end{aligned}
\end{equation}

where \( A, B, C, D \sim \text{Uniform}(0, 1) \); and \( a, b, c, d, e, f, g, h \sim \{2\pi, 4\pi\} \).

Details on the architecture choices for the RL framework and the hyperparameter can be found in the Appendix \ref{sec:numerical_parameters} and \ref{app:AD}. Moreover, visualization of training metrics are displayed in Appendix \ref{sec:2D_Advection_Training_Metrics}.

%\subsection{2D Advection Equation}
\input{advection}

\section{Conclusions}
\label{sec:conclusions}
We present a novel approach for acquiring and deploying synthetic data to learn the closures of under-resolved PDEs through reinforcement learning. The synthetic training data is generated using the MMS. This approach is a cost-effective method to generate training data and mitigates the computational burden that would be associated with high-accuracy numerical simulations. %The empirical results demonstrate the effectiveness of the learned closure model in generalizing well to unseen test data, underscoring its potential as a viable alternative to traditional, computationally intensive training methodologies.

We validated our approach across three distinct numerical illustrations: the 1D Burgers’ equation, the 2D Burgers’ equation, and the 2D advection equation. The results consistently demonstrate the capability of RL-based closures to accurately enhance coarse-grid simulations. Specifically, the learned closures are able to reduce simulation errors substantially in both in-distribution (inhomogeneous PDEs) and out-of-distribution (homogeneous PDEs) settings, achieving significant median error reductions. Moreover, the RL-based closures showed superior generalization capabilities compared to a FNO-based approach. Our results highlight the potential for reducing both time and energy costs in data-driven closure modeling, making such models more accessible for practical applications.

Despite these promising results, certain limitations must be acknowledged. To generate synthetic training data, complete knowledge of the underlying PDE is required. As a consequence, our method is not directly applicable to systems with partially unknown dynamics or uncertain governing equations. Additionally, the computational efficiency gains demonstrated here, while significant, should be further evaluated against more complex, higher-dimensional PDE systems.

Future research will focus on a comprehensive sensitivity analysis of PDE parameters and the optimal selection of MMS solutions. Although MMS allows the generation of synthetic training data for arbitrary proposed solutions, we believe that a deeper investigation is warranted to determine if there is an optimal set of solutions that yield particularly robust training outcomes and strong extrapolative capabilities. Additionally, integrating our approach to synthetic data generation with differentiable solvers presents an exciting promising direction, potentially enabling the training of alternative data-driven closure models and further broadening the applicability of our methodology.

% Only for non-anonymized version !
\section*{Acknowledgements}
S.K. and P.K. acknowledge support by the Defense Advanced Research Projects Agency (DARPA)
through Award HR00112490489. F.A. acknowledges support by the Applied Mathematics (M2dt MMICC center) activity within the U.S. Department of Energy, Office of Science, Advanced Scientific Computing Research, under Contract DE-AC02-06CH11357.

%\section*{References}
\bibliographystyle{plainnat}
\bibliography{bibliography}

%%%%%%%%%%%%%%%%%%%%%%%%%%%%%%%%%%%%%%%%%%%%%%%%%%%%%%%%%%%%

\appendix

\include{appendix}

%%%%%%%%%%%%%%%%%%%%%%%%%%%%%%%%%%%%%%%%%%%%%%%%%%%%%%%%%%%%

\end{document}

%% file: 1dBurger.tex
\subsubsection{In-Distribution Predictions}
\begin{figure}[h]
    \centering
    \includegraphics[width=1.0\linewidth]{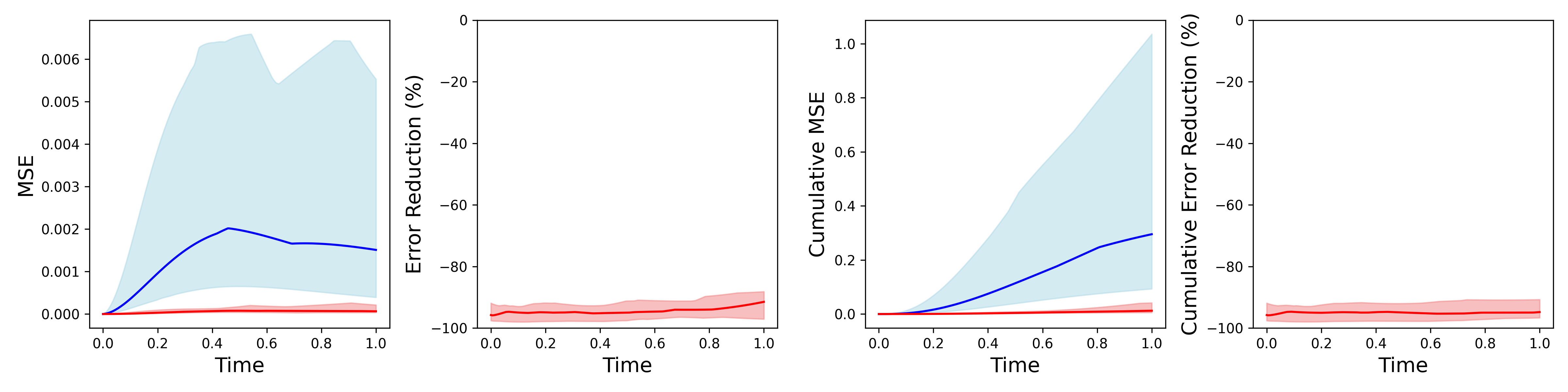}
   \caption{Results for the inhomogeneous 1D Burgers' equation. The mean squared error (MSE) is calculated with respect to the proposed analytical MMS solution. The plots show median values for the CGS (\textcolor{CGS}{\rule[0.5ex]{1.em}{1.5pt}}) and the Closure-RL (\textcolor{RL}{\rule[0.5ex]{1.em}{1.5pt}}) across 30 different MMS solutions. The shaded regions around the medians represent the interquartile range (25th to 75th percentile).}
    \label{fig:1D_Burgers_Forced_Stats}
\end{figure}

We used the learned closure model to predict the temporal evolution of the solution to an inhomogeneous Burgers' equation. The right-hand side used are based on 30 unseen samples of equation \ref{eq:mms_1D_Burgers}.
Figure \ref{fig:1D_Burgers_Forced_Stats} shows the evolution of the mean squared error (MSE) over the duration of the CG and Closure-RL simulations evaluated on 30 samples of equation \ref{eq:mms_1D_Burgers}. The results indicate that the median error reduction (with small interquartile range) is greater than $80\%$ for all times during the simulation. The cumulative error plot, which is obtained by adding up all MSE values until the current time, confirms this and shows that for the entire duration of the simulation there is a median error reduction of close to $100\%$. Based on these results, we conclude that Closure-RL learned to significantly reduce the MSE caused by the CGS for the family of inhomogeneous PDEs it was trained on.

\begin{figure}[h]
    \centering
    \includegraphics[width=1.0\linewidth]{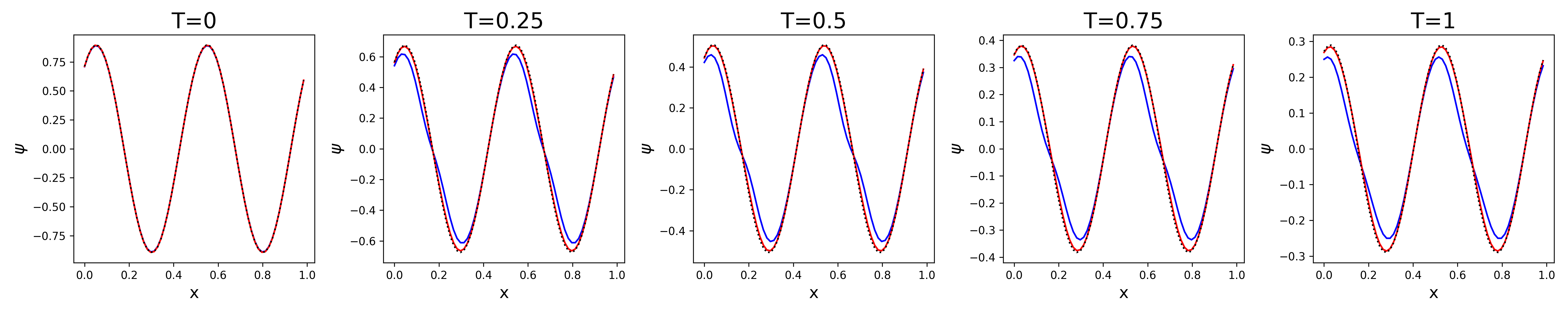}
    \caption{Visualization of the evolution of $\psi$ for the inhomogeneous 1D Burgers' equation at five different time snapshots. The figure shows the evolution for the CGS (\textcolor{CGS}{\rule[0.5ex]{1.em}{1.5pt}}), the Closure-RL simulation (\textcolor{RL}{\rule[0.5ex]{1.em}{1.5pt}}), and the MMS solution ($\cdots$).}
    \label{fig:1D_Burgers_Forced_0}
\end{figure}

An illustrative sample simulation is shown in figure \ref{fig:1D_Burgers_Forced_0}. The Closure-RL simulation is very accurate compared to the CG simulation for all five snapshots shown. Closure-RL learns an accurate correction of both the magnitude and the phase of the CGS.
\subsubsection{Out-of-Distribution Predictions}
\begin{figure}[h]
    \centering
    \includegraphics[width=1.0\linewidth]{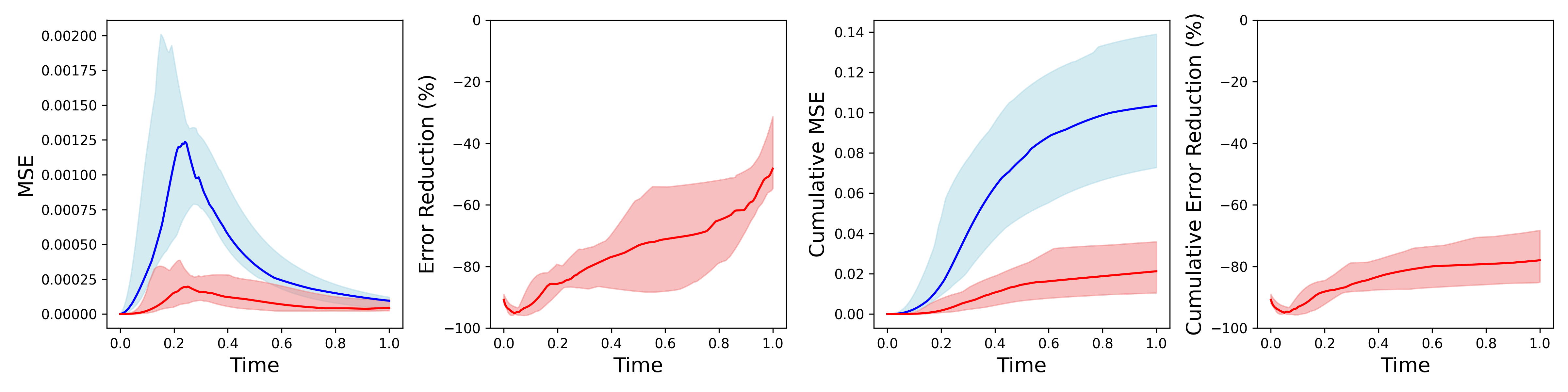}
    \caption{Results for the homogeneous 1D Burgers' equation. The MSE is calculated with respect to the FGS. The plots show median values for the CGS (\textcolor{CGS}{\rule[0.5ex]{1.em}{1.5pt}}) and the Closure-RL (\textcolor{RL}{\rule[0.5ex]{1.em}{1.5pt}}) across 30 different MMS solutions. The shaded regions around the medians represent the interquartile range (25th to 75th percentile).}
    \label{fig:1D_Burgers_Unforced_Stats}
\end{figure}

We predicted the solution to the homogeneous PDE with the learned closure term. The results are shown in figure \ref{fig:1D_Burgers_Unforced_Stats}. Even though Closure-RL had not seen an homogeneous PDE during training, it is able to consistently reduce the error over the duration of the simulation achieving a median cumulative error reduction of approximately $80\%$ over the entire simulation. We note that this error reduction is lower than for the in-distribution predictions. However, this is expected as this extrapolative prediction task is in general more challenging. 

\begin{figure}[h]
    \centering
    \includegraphics[width=1.0\linewidth]{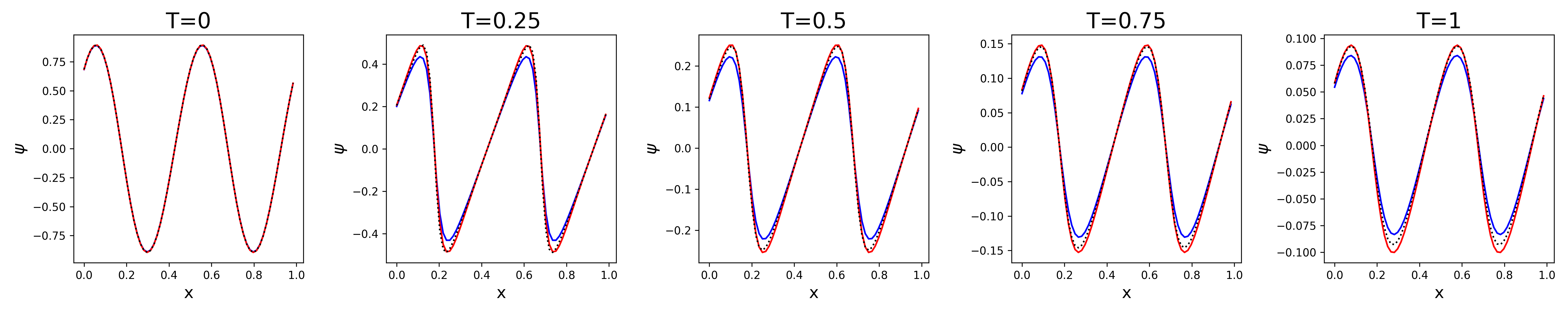}
    \caption{Visualization of the evolution of $\psi$ for the homogeneous 1D Burgers' equation at five different time snapshots. The figure shows the evolution for the CG (\textcolor{CGS}{\rule[0.5ex]{1.em}{1.5pt}}), the Closure-RL (\textcolor{RL}{\rule[0.5ex]{1.em}{1.5pt}}), and the FG ($\cdots$) simulations.}
    \label{fig:1D_Burgers_Unforced_0}
\end{figure}

In the illustrative sample for the unforced 1D Burgers' equation shown in figure \ref{fig:1D_Burgers_Unforced_0} the CGS's main source of error stems from the magnitude in the extrema not being sufficiently large. The phase on the other hand is accurate. Hence, the ideal Closure-RL model should account for the exceedingly large diffusion introduced through the coarsening of the grid. For the chosen sample, we can see that at $T=0.25$ and $T=0.5$ the Closure-RL simulation agrees almost perfectly with the FGS. In contrast, for $T=0.75$ and $T=1$ the magnitude of the negative extrema are captured less accurately over time albeit still significantly better than in the CGS.

Moreover, we compared the predictions obtained using reinforcement learning closures with the predictions of a neural operator in the Appendix \ref{app:fno}. While the results indicate only minimal differences for in-distribution tasks, the RL closures significantly outperform the neural operator based approach for the out-of-distribution predictions. Both models were trained using the same synthetic data generated by the MMS. This indicates that our synthetic data generation approach can easily be used to train other models as well, but for the specific task of accurately predicting coarse-grained PDEs using RL based closures is a strong model choice.

%% file: 2dBurger.tex
\subsubsection{In-Distribution-Predictions}
\begin{figure}[h]
    \centering
    \includegraphics[width=1.0\linewidth]{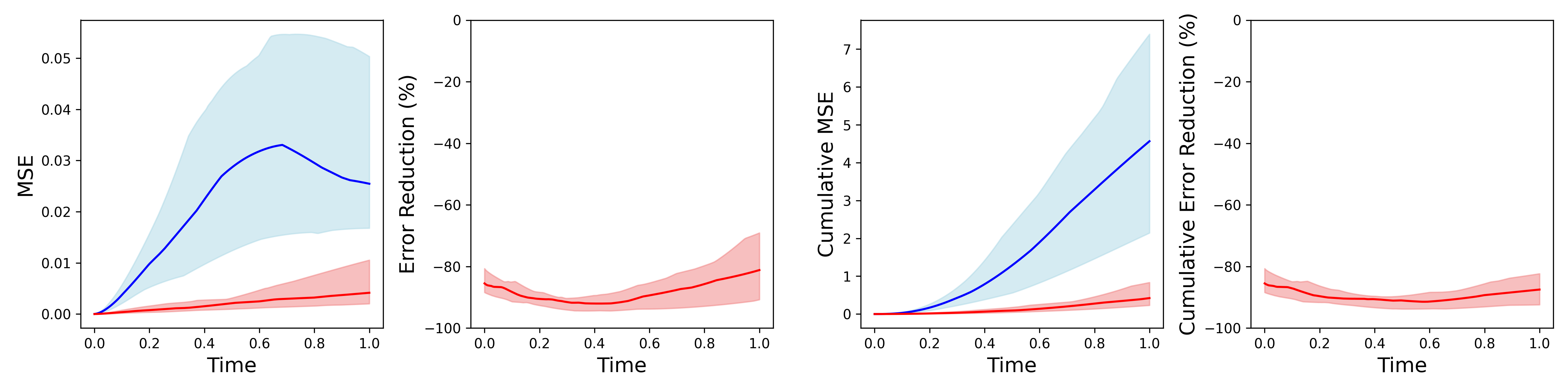}
    \caption{Results for the inhomogeneous 2D Burgers' equation. The MSE is calculated with respect to the analytical MMS. The plots show median values for the CGS (\textcolor{CGS}{\rule[0.5ex]{1.em}{1.5pt}}) and the Closure-RL (\textcolor{RL}{\rule[0.5ex]{1.em}{1.5pt}}) across 30 different MMS solutions. The shaded regions around the medians represent the interquartile range (25th to 75th percentile).} 
    \label{fig:2D_Burgers_Forced_Stats}
\end{figure}

Figure \ref{fig:2D_Burgers_Forced_Stats} shows the results for the in-distribution predictions for the 2D Burgers' equation. In comparison to the results for the inhomogeneous 1D Burgers' equation the MSEs for the 2D Burgers' equation are an order of magnitude larger. This is due to the grid spacing being twice as large as for the 1D Burgers' equation, resulting in a worse approximation and hence a larger error of the CGS. The median and all values within the interquartile range of error reduction are greater than $60\%$ during all simulation times. There also is no overlap between the interquartile ranges of the CG and Closure-RL simulations for the MSE plot. Therefore, the Closure-RL consistently outperforms the CGS for different inhomogeneous 2D Burger equations. 
An illustrative sample solution is displayed in figure \ref{fig:2D_Burgers_Forced_0} in the Appendix \ref{Sec:vis2D}.
 %The trained closure model maintains the structure of the solution more accurately than the CGS, however, the structure is still slightly smoother than in the MMS solution. 

\subsubsection{Out-of-Distribution Predictions}
\begin{figure}[h]
    \centering
    \includegraphics[width=1.0\linewidth]{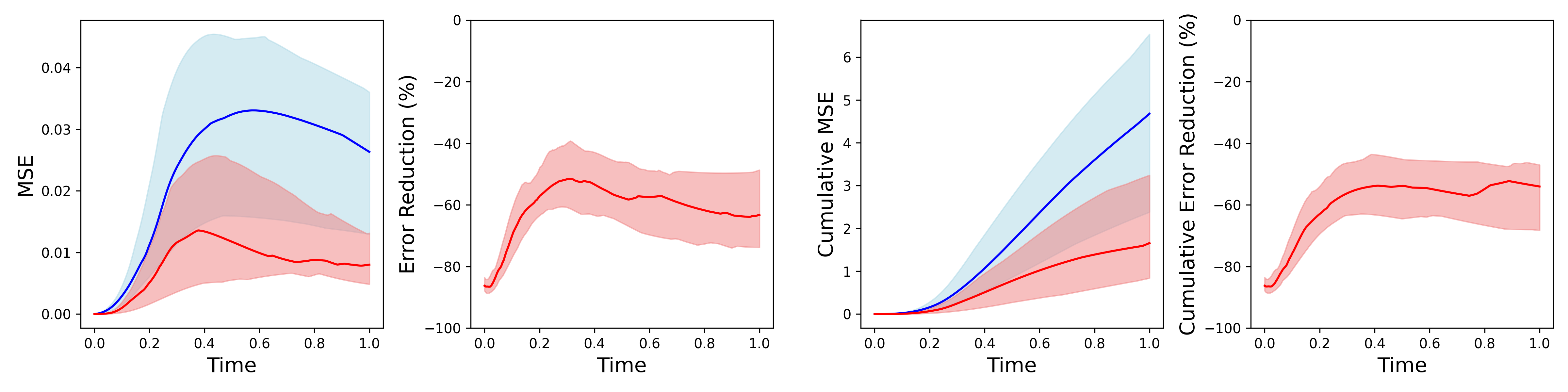}
    \caption{Results for the homogeneous 2D Burgers' equation. The MSE is calculated with respect to the FGS. The plots show median values for the CGS (\textcolor{CGS}{\rule[0.5ex]{1.em}{1.5pt}}) and the Closure-RL (\textcolor{RL}{\rule[0.5ex]{1.em}{1.5pt}}) across 30 different MMS solutions. The shaded regions around the medians represent the interquartile range (25th to 75th percentile).}
    \label{fig:2D_Burgers_Unforced_Stats}
\end{figure}

In figure \ref{fig:2D_Burgers_Unforced_Stats} we can see that Closure-RL generalizes to the homogeneous 2D Burgers' equation. The cumulative error reduction for the entire simulation is greater than $40\%$ for all times of the simulation. Unlike for the homogeneous 1D Burgers' equation, the error reduction of Closure-RL does not worsen towards the end of the simulation but improves again, reaching a median $60\%$ by the end.
An illustrative sample solution is displayed in figure \ref{fig:2D_Burgers_Unforced_0} in the Appendix \ref{Sec:vis2D}.

 %For this sample, the CGS has insufficient magnitude, especially in the extrema throughout the simulation. Furthermore,the structure of the solution is lost. At $T=1$ the extrema of the CGS solution are isolated from other large magnitude zones through a low magnitude line encircling them. In contrast, in the FGS the extrema are not encircled by low magnitude zones, instead the low magnitude zones cross the solution domain diagonally. The Closure-RL solution at $T=1$ is both able to significantly increase the accuracy of the magnitude of the solution and correct the structural mistakes occurring in the CGS.

%% file: advection.tex
\subsubsection{In-Distribution Predictions}
\begin{figure}[h]
    \centering
    \includegraphics[width=1.0\linewidth]{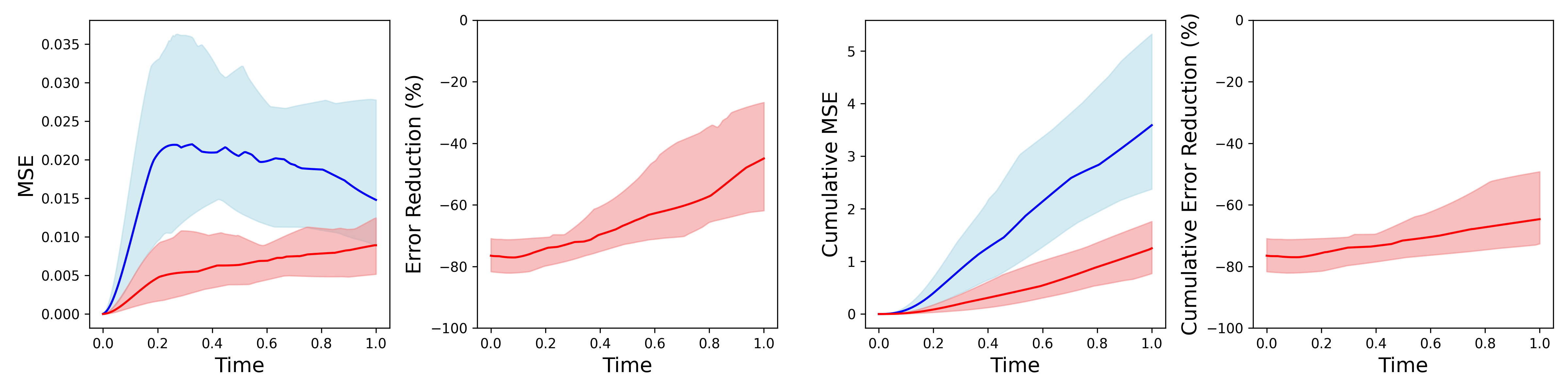}
    \caption{Results for the inhomogeneous 2D advection equation. The MSE is calculated with respect to the analytical MMS. The plots show median values for the CGS (\textcolor{CGS}{\rule[0.5ex]{1.em}{1.5pt}}) and the Closure-RL (\textcolor{RL}{\rule[0.5ex]{1.em}{1.5pt}}) across 30 different MMS solutions. The shaded regions around the medians represent the interquartile range (25th to 75th percentile).}
    \label{fig:2D_Advection_Forced_Stats}
\end{figure}

The in-distribution results for the 2D advection equation are shown in figure \ref{fig:2D_Advection_Forced_Stats}. We observe a median error reduction between $80\%$ and $40\%$. Additionally, for almost the entire simulation time the interquartile ranges of the CG and Closure-RL simulation's MSEs do not overlap, providing strong evidence for the effectiveness of RL Closures trained with synthetic data for a diverse amount of test trajectories.
An illustrative sample prediction for the inhomogeneous 2D advection equation in figure \ref{fig:2D_Advection_Forced} can be found in the Appendix \ref{sec:visAD}.

% the structure is kept almost perfectly by the CGS. Only the magnitude is not accurate in parts. If we examine the Figure at $T=1$ again the top part of the solution for the CGS has an extrema on the left and right with insufficiently large magnitude. Closure-RL visibly improves this, leading to a higher accuracy.

\subsubsection{Out-of-Distribution Predictions}
\begin{figure}[h]
    \centering
    \includegraphics[width=1.0\linewidth]{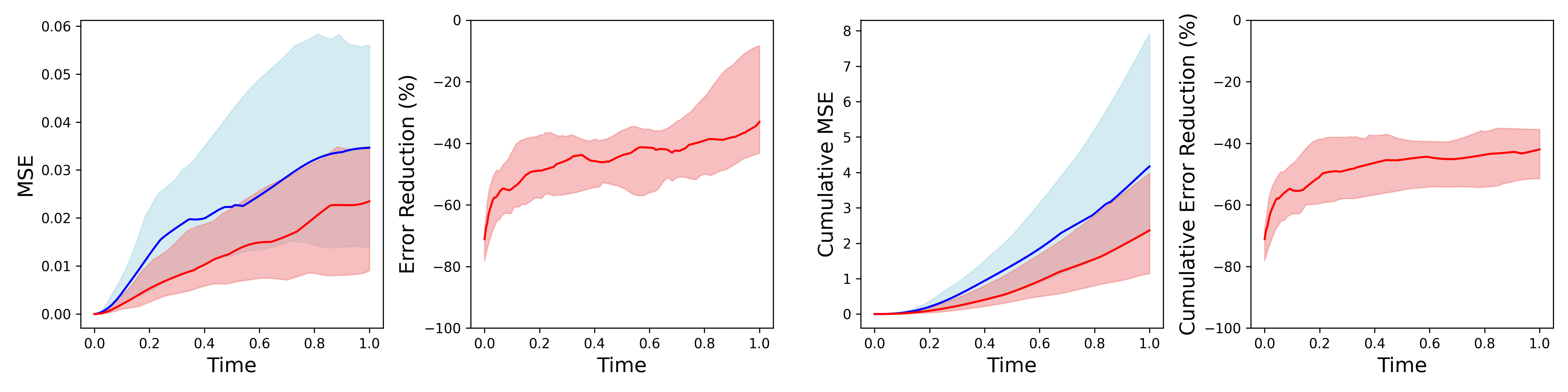}
    \caption{Results for the unforced 2D advection equation. The MSE is calculated with respect to the FGS. The plots show median values for the CGS (\textcolor{CGS}{\rule[0.5ex]{1.em}{1.5pt}}) and the Closure-RL (\textcolor{RL}{\rule[0.5ex]{1.em}{1.5pt}}) across 30 different MMS solutions. The shaded regions around the medians represent the interquartile range (25th to 75th percentile).}
    \label{fig:2D_Advection_Unforced_Stats}
\end{figure}

The out of distribution results are shown in figure \ref{fig:2D_Advection_Unforced_Stats}. The results indicate a median error reduction over the duration of the episode of more than $40\%$ and thus a significant improvement compared to the CGS. We note however, that there is a large overlap of the upper interquartile range of the Closure-RL simulation and the lower interquartile range of the CGS simulation. This indicate that for this example the error reduction by Closure-RL is less consistent compared to our other two test cases. 
An illustrative sample prediction for the homogeneous 2D advection can be found in figure \ref{fig:2D_Advection_Unforced} in the Appendix \ref{sec:visAD}.

%equation the majority of the error arises from insufficient magnitudes of the solution. Closure-RL is able to correct this mostly. Comparing the three simulation at $T=0.75$, the dark blue region on the left of the solution domain is not captured perfectly by Closure-RL and even less sufficiently by the CGS when comparing to the FGS.

%% file: appendix.tex
\newpage
\section{Training Details and Hyperparameters}
\label{sec:numerical_parameters}
This section contains the training details and hyperparameters employed for all three test cases. Moreover, we report parameters for the CGS and FGS here. As we do not require any FGS data for training, these values are only relevant to compute the test errors for homogeneous cases. All values without a tilde are associated with the FGS.

We used a similar architecture for both the neural network that encodes the policy and the neural network that encodes the value function used for the PPO. In fact, both share the same backbone as suggested in \cite{zhang2017learningdeepcnndenoiser}. After 6 convolutional layers (Conv2D\_1 - Conv2D\_6) process the input, the resulting outputs are processed by either a convolutional layer Conv2D\_{\(\pi\) to compute the mean and standard deviation for the action predictions or another convolutional layer Conv2D\_{\(\mathcal{V}\)} to compute the value function. Between each convolutional layer, we employ ReLU activation functions. Variations in the number of layers and layer parameters were also explored  and did not impact significantly the performance.

\begin{table}[h]
\centering
\begin{tabular}{|c c c c c c|} 
 \hline
 Layer & Input Channels & Output Channels & Kernel Size & Padding & Dilation  \\ [0.5ex] 
 \hline\hline
 Conv2D\_1 & 3, 6, 5 & 64 & 3 & 1 & 1  \\ 
 \hline
 Conv2D\_2 & 64 & 64 & 3 & 2 & 2  \\ 
 \hline
 Conv2D\_3 & 64 & 64 & 3 & 3 & 3  \\ 
 \hline
 Conv2D\_4 & 64 & 64 & 3 & 4 & 4  \\ 
 \hline
 Conv2D\_5 & 64 & 64 & 3 & 3 & 3  \\ 
 \hline
 Conv2D\_6 & 64 & 64 & 3 & 2 & 2  \\ 
 \hline
 Conv2D\_{\(\pi\)} & 64 & 2, 4, 2 & 3 & 1 & 1  \\ 
 \hline
 Conv2D\_{\(\mathcal{V}\)} & 64 & 1 & 3 & 1 & 1  \\ [1ex] 
 \hline
\end{tabular}
\caption{Table detailing the architecture of the  neural network. The Conv2D\_1 and Conv2D\_{\(\pi\)} layers have multiple entries for the number of input/output channels. The values represent in order the value used in the 1D Burgers', 2D Burgers', and advection equation.}
\label{fig:table_receptive_field}
\end{table}

For all three numerical illustrations, the RL policy is trained for 1000 epochs with each epoch consisting of multiple episodes, i.e time series of PDE data.  An episode ends after either 200 time steps are complete or the mean absolute error (MAE) w.r.t. the MMS solution exceeds a predefined threshold. The latter ensures that there are no blow-ups due to erroneous actions of the RL model during training which would cause the training process to end prematurely. During this training the MMS parameters are sampled from the set of potential parameters defined for each PDE.\\\\
At the end of each epoch, the RL model is evaluated on a validation set consisting of 32 different initial conditions of the homogeneous PDE. This validation set is employed to additionally track the accuracy of the learned closure model for the homogeneous test case and allows us to choose a closure model that is the best choice with regards to performance on both homogeneous and inhomogeneous PDEs. We note that all test sets employed are different from the validation sets and contain unseen initial conditions, PDE parameters and forcing terms.

\begin{table}[h!]
\begin{center}
\begin{tabular}{ c c }
\hline
 Learning Rate & $10^{-5}$ \\ 
 Entropy Coefficient & $0.02$  \\  
 Discount Factor & $1$  \\
 Epoch & $1000$ \\
 Transitions per Epoch & $2500$ \\
 Episodes Per Policy Update & $10$ \\
 Batch Size & $50$ \\
 Repeat per Collect & $2$ \\
 Validation Episodes per Epoch & $32$ \\
 Padding & Circular \\
\hline
\end{tabular}
\caption{RL Parameters for all three experiments}
\label{table:rl_parameters}
\end{center}  % End centering
\end{table}

\subsection{1D Burgers' Equation}
\label{app:1D}
Additional Parameters specific to the 1D Burgers' equation example are provided in Table \ref{table:parameters_burgers_1D}. The RL framework was trained for 1.3 hours on a single NVIDIA A100 GPU and used 1.8 GB of GPU memory. Moreover, some supportive task executed on CPU (Intel Ice Lake CPUs) for processing the environments that feed data to the GPU for model updates. 
\begin{table}[h!]
\begin{center}
\begin{tabular}{ c c }
\hline
 $\Omega$ & $[0,1]$ \\ 
 $\Tilde{N}$ & $64$  \\  
 $N$ & $2048$  \\
 $\Tilde{\Delta x}$ & $\approx 1.56\times 10^{-2}$ \\
 $\Delta x$ & $\approx 4.88\times 10^{-4}$ \\
 $\Tilde{\Delta t}$ & $ 5\times 10^{-3}$ \\
 $\Delta t$ & $10^{-5}$ \\
 $\nu$ & $10^{-2}$ \\
 MAE Threshold & $2\times10^{-2}$ \\
\hline
\end{tabular}
\caption{Parameter Burgers' 1D}
\label{table:parameters_burgers_1D}
\end{center}  % End centering
\end{table}
\subsection{2D Burgers' Equation}
\label{app:2D}
Additional Parameters specific to the 2D Burgers' equation example are provided in Table \ref{table:parameters_burgers_2D}. The RL framework was trained for 2.4 hours on a single NVIDIA A100 GPU and used 18.0 GB of GPU memory. Moreover, some supportive task executed on CPU (Intel Ice Lake CPUs) for processing the environments that feed data to the GPU for model updates. 
\begin{table}[h!]
\begin{center}
\begin{tabular}{ c c }
\hline
 $\Omega$ & $[0,1]\times [0,1]$ \\ 
 $\Tilde{N_x}$, $\Tilde{N_y}$  & $32$  \\  
 $N_x$, $N_y$ & $128$  \\
 $\Tilde{\Delta x}$, $\Tilde{\Delta y}$ & $\approx 3.13\times 10^{-2}$ \\
 $\Delta x$, $\Delta y$ & $\approx 7.81\times 10^{-3}$ \\
 $\Tilde{\Delta t}$ & $ 5\times 10^{-3}$ \\
 $\Delta t$ & $10^{-4}$ \\
 $\nu$ & $5\times 10^{-3}$ \\
  MAE Threshold & $10^{-1}$ \\
\hline
\end{tabular}
\caption{Parameter Burgers' 2D}
\label{table:parameters_burgers_2D}
\end{center}  % End centering
\end{table}
\subsection{2D Advection Equation}
\label{app:AD}
Additional Parameters specific to the 2D advection equation are provided in Table \ref{table:parameters_advection_2D}.  The RL framework was trained for 1.6 hours on a single NVIDIA A100 GPU and used 14.0 GB of GPU memory. Moreover, some supportive task executed on CPU (Intel Ice Lake CPUs) for processing the environments that feed data to the GPU for model updates. 
\begin{table}[h!]
\begin{center}
\begin{tabular}{ c c }
\hline
 $\Omega$ & $[0,1]\times [0,1]$ \\ 
 $\Tilde{N_x}$, $\Tilde{N_y}$  & $32$  \\  
 $N_x$, $N_y$ & $128$  \\
 $\Tilde{\Delta x}$, $\Tilde{\Delta y}$ & $\approx 3.13\times 10^{-2}$ \\
 $\Delta x$, $\Delta y$ & $\approx 7.81\times 10^{-3}$ \\
 $\Tilde{\Delta t}$ & $ 5\times 10^{-3}$ \\
 $\Delta t$ & $10^{-4}$ \\
  MAE Threshold & $5\times10^{-2}$ \\
\hline
\end{tabular}
\caption{Parameter Advection 2D}
\label{table:parameters_advection_2D}
\end{center}  % End centering
\end{table}

\newpage
\section{Training Metrics}

\subsection{1D Burgers' Equation Training Metrics}
\label{sec:tr1D}
The Reinforcement Learning framework is trained over 1000 epochs in total using the synthetic training data only. In each epoch, 2500 CGS transitions are evaluated - distributed in episodes of varying length.
In Figure \ref{fig:1D_Burgers_train_metrics}, the evolution of the reward and episode length achieved during the training of Closure-RL on the 1D Burgers' equation is visualized. At the start of training, the random actions of Closure-RL cause the episodes to be terminated prematurely as the solution diverges to far from the ground truth solution, i.e. the FGS. The episodes are automatically truncated if the error of the CGS is larger than $2\times10^{-1}$ MAE compared to the synthetic MMS solution at any time step. Only after the policy is trained for a number of epochs does the moving average of the episode length reach its maximum of 200. Subsequently,  Closure-RL has learned to correct the error of the CGS such that the simulation stays close to the FGS.  The reward plot indicates  a similar trend. The reward plateaus after 200 epochs with a positive value indicating an error reduction over an average episode. 

\begin{figure}[h]
    \centering
    \includegraphics[width=1.0\linewidth]{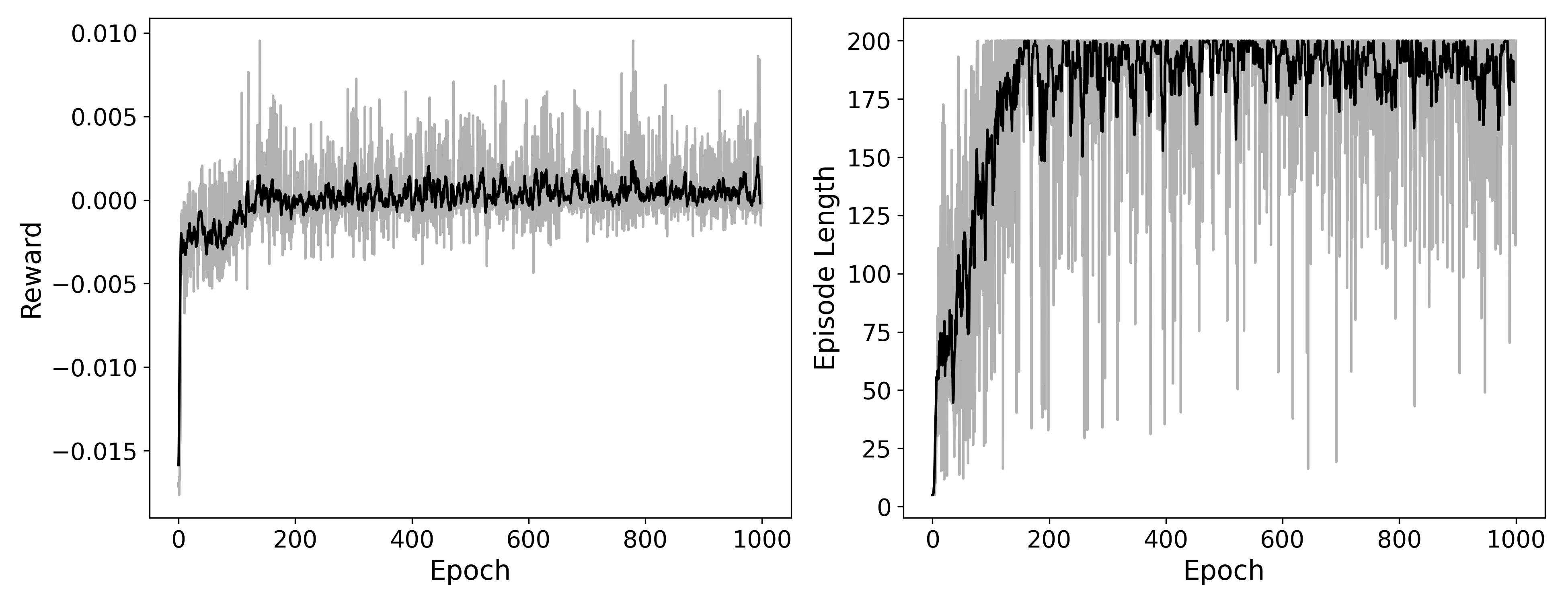}
   \caption{Illustration of training metrics recorded for the training of Closure-RL on the 1D Burgers' equation. The grey line represents the values recorded. The black line represents a moving average of the values with window size 10. Left: The average reward achieved during an episode. Right: The average episode length achieved during an episode.}
    \label{fig:1D_Burgers_train_metrics}
\end{figure}

\subsection{2D Burgers' Equation Training Metrics}
\label{sec:2D_Burgers_Training_Metrics}
\begin{figure}[h!]
    \centering
    \includegraphics[width=1.0\linewidth]{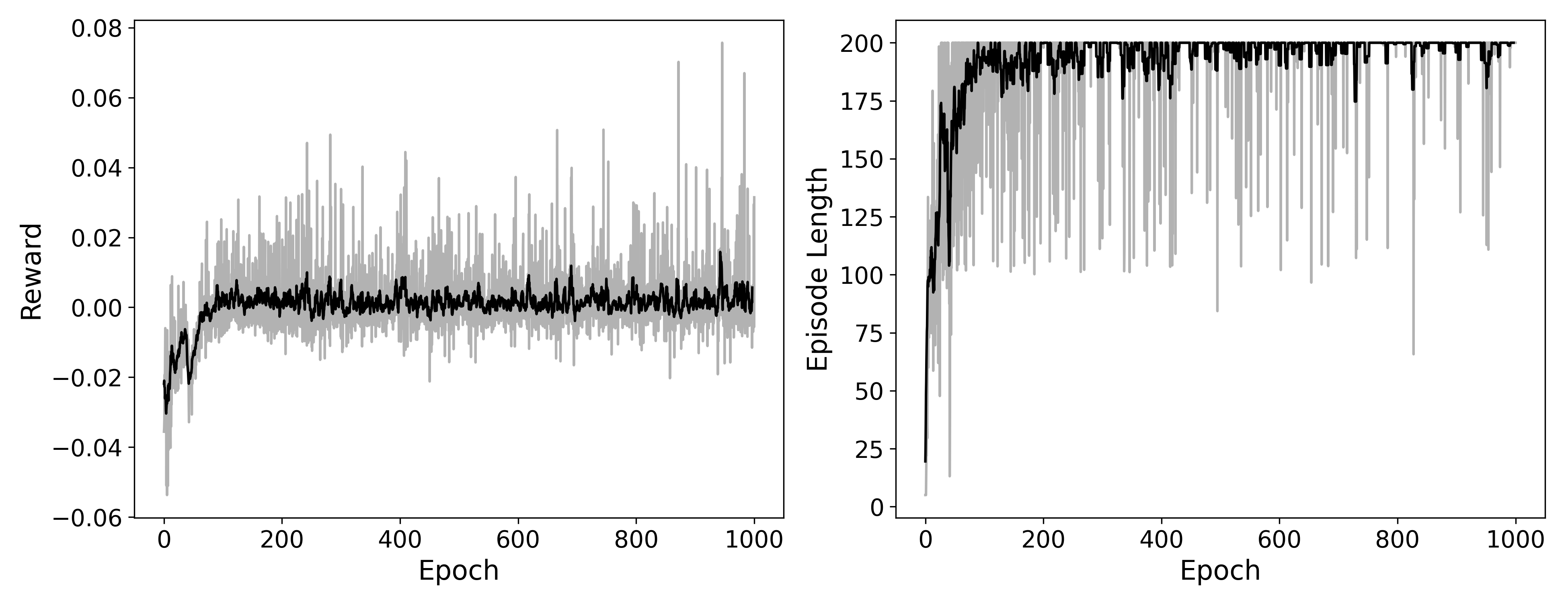}
   \caption{Illustration of training metrics recorded for the training of Closure-RL on the 2D Burgers' equation. The grey line represents the values recorded. The black line represents a moving average of the values with window size 10. Left: The average reward achieved during an episode. Right: The average episode length achieved during an episode.}
    \label{fig:2D_Burgers_train_metrics}
\end{figure}
\newpage
\subsection{2D Advection Equation Training Metrics}
\label{sec:2D_Advection_Training_Metrics}
\begin{figure}[h!]
    \centering
    \includegraphics[width=1.0\linewidth]{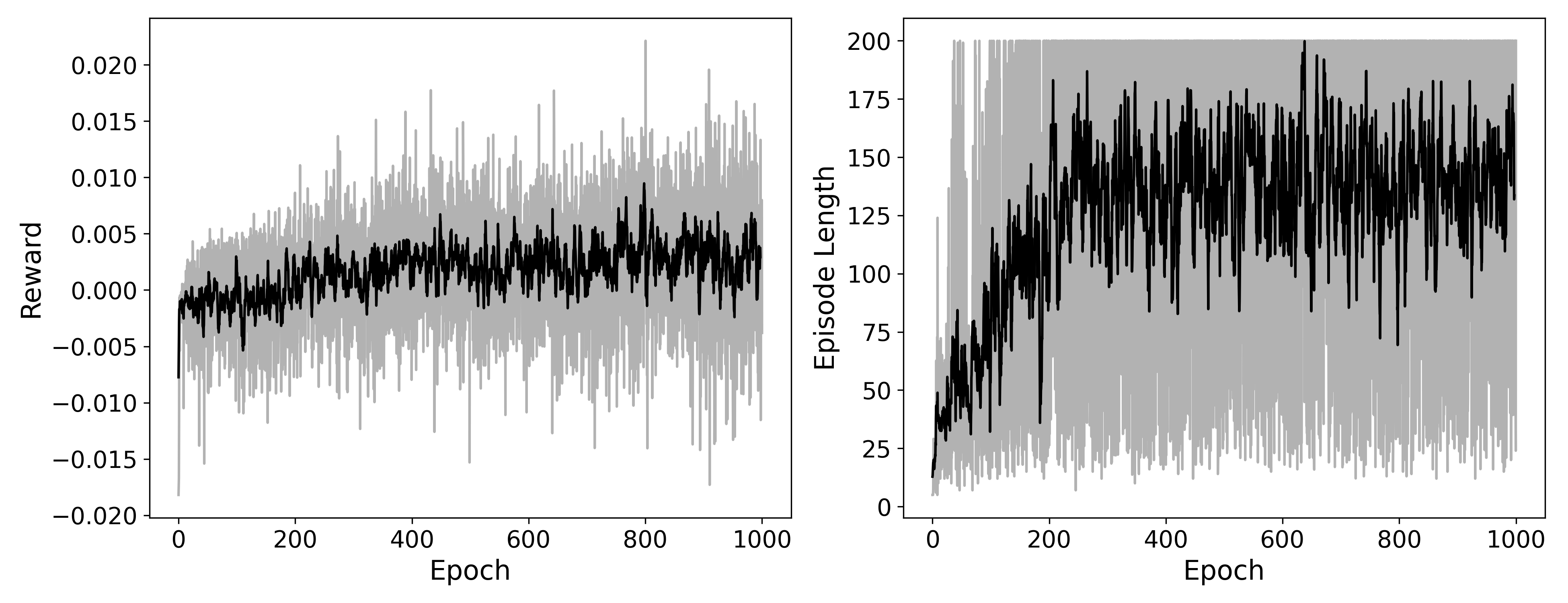}
   \caption{Illustration of training metrics recorded for the training of Closure-RL on the 2D advection equation. The grey line represents the values recorded. The black line represents a moving average of the values with window size 10. Left: The average reward achieved during an episode. Right: The average episode length achieved during an episode.}
    \label{fig:2D_advection_train_metrics}
\end{figure}

\newpage
\section{Additional Visualizations }

\subsection{2D Burgers' Equation}
\label{Sec:vis2D}
\begin{figure}[h]
    \centering
    \includegraphics[width=1.0\linewidth]{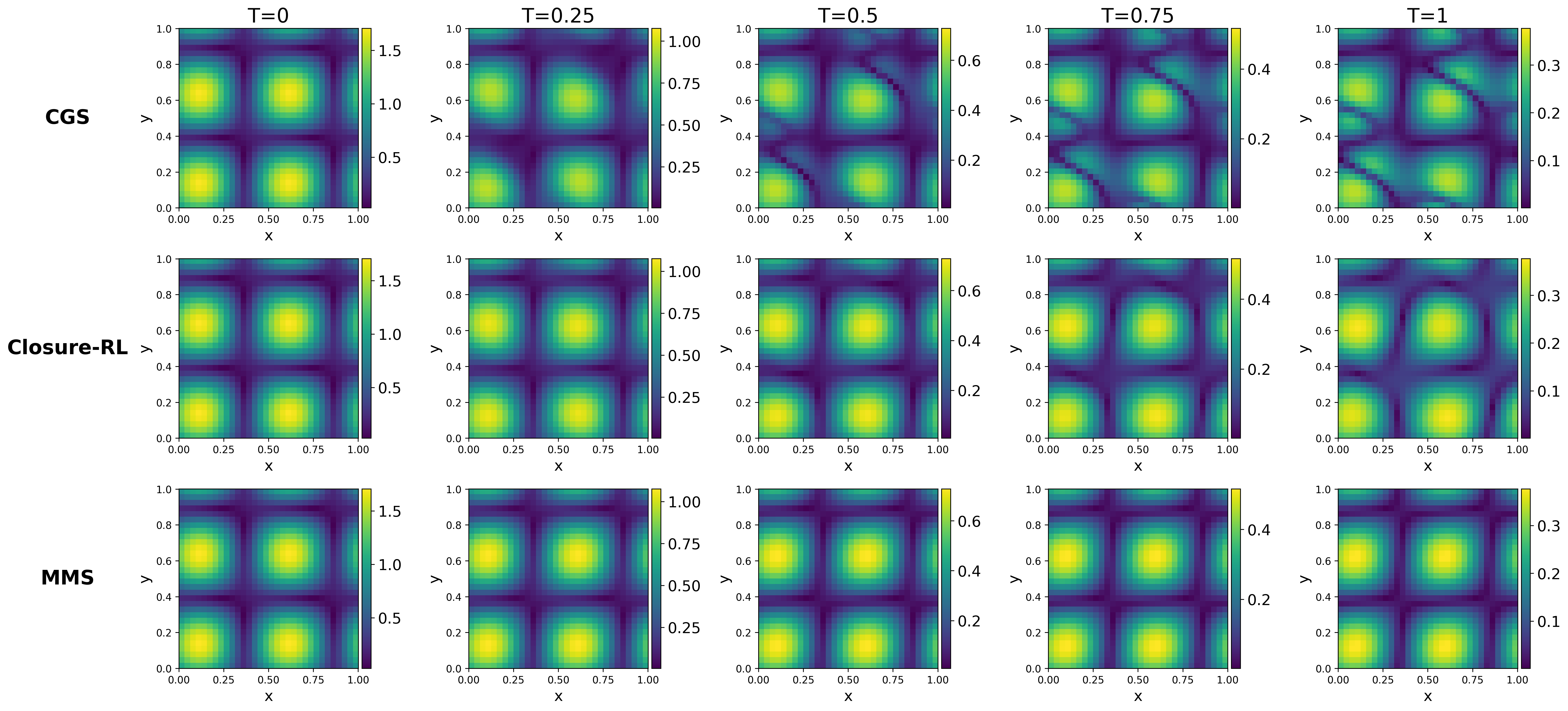}
    \caption{Visualization of the evolution of the velocity magnitude of $\psi$ for the inhomogeneous 2D Burgers' equation at five different time snapshots. The velocity magnitude is calclated as \(\sqrt{u_{i,j}^2+v_{i,j}^2}\). The trained closure model maintains the structure of the solution more accurately than the CGS, however, the structure is still slightly smoother than in the MMS solution. }
    \label{fig:2D_Burgers_Forced_0}
\end{figure}

\begin{figure}[h]
    \centering
    \includegraphics[width=1.0\linewidth]{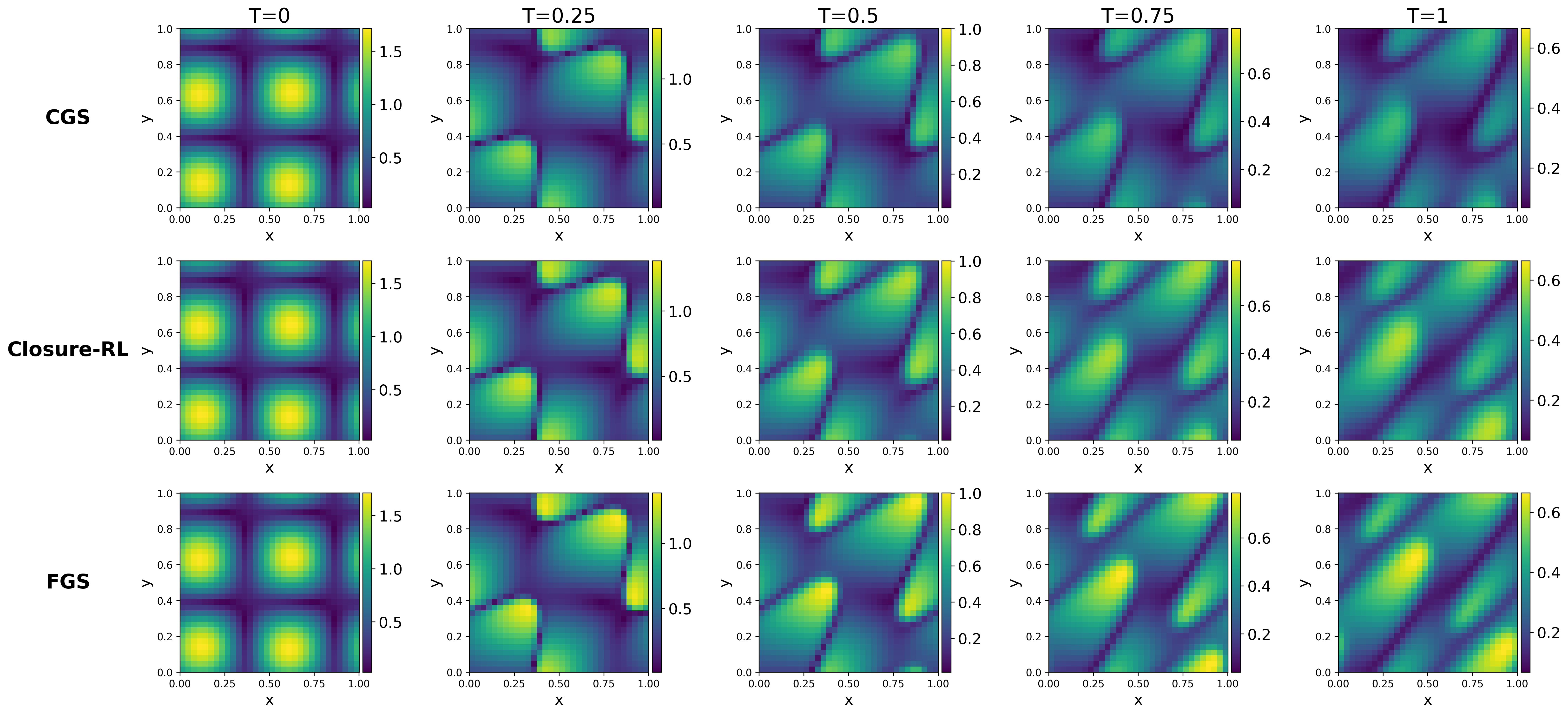}
    \caption{Visualization of the evolution of the velocity magnitude of $\psi$ for the homogeneous 2D Burgers' equation at five different time snapshots. The CGS has insufficient magnitude, especially in the extrema throughout the simulation. Furthermore,the structure of the solution is lost. At $T=1$ the extrema of the CGS solution are isolated from other large magnitude zones through a low magnitude line encircling them. In contrast, in the FGS the extrema are not encircled by low magnitude zones, instead the low magnitude zones cross the solution domain diagonally. The Closure-RL solution at $T=1$ is both able to significantly increase the accuracy of the magnitude of the solution and correct the structural mistakes occurring in the CGS.}
    \label{fig:2D_Burgers_Unforced_0}
\end{figure}

\newpage
\subsection{2D Advection Equation}
\label{sec:visAD}
\begin{figure}[h]
    \centering
    \includegraphics[width=1.0\linewidth]{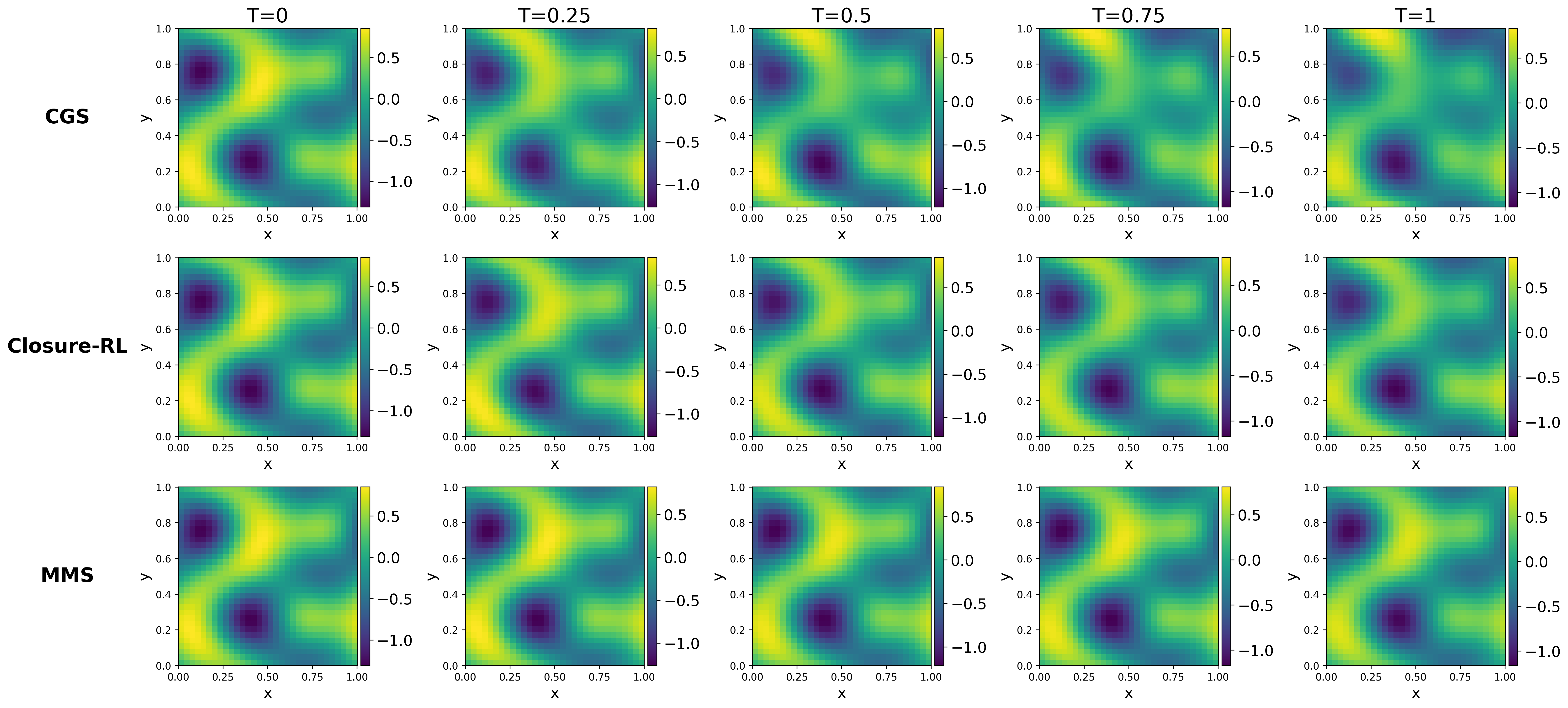}
    \caption{Visualization of the evolution of $\psi$ for the inhomogeneous 2D advection equation at five different time snapshots. The structure of the solution is kept almost perfectly by the CGS. Only the magnitude is not accurate in parts. If we examine the figure at $T=1$ again the top part of the solution for the CGS has an extrema on the left and right with insufficiently large magnitude. Closure-RL visibly improves this, leading to a higher accuracy.}
    \label{fig:2D_Advection_Forced}
\end{figure}

\begin{figure}[h]
    \centering
    \includegraphics[width=1.0\linewidth]{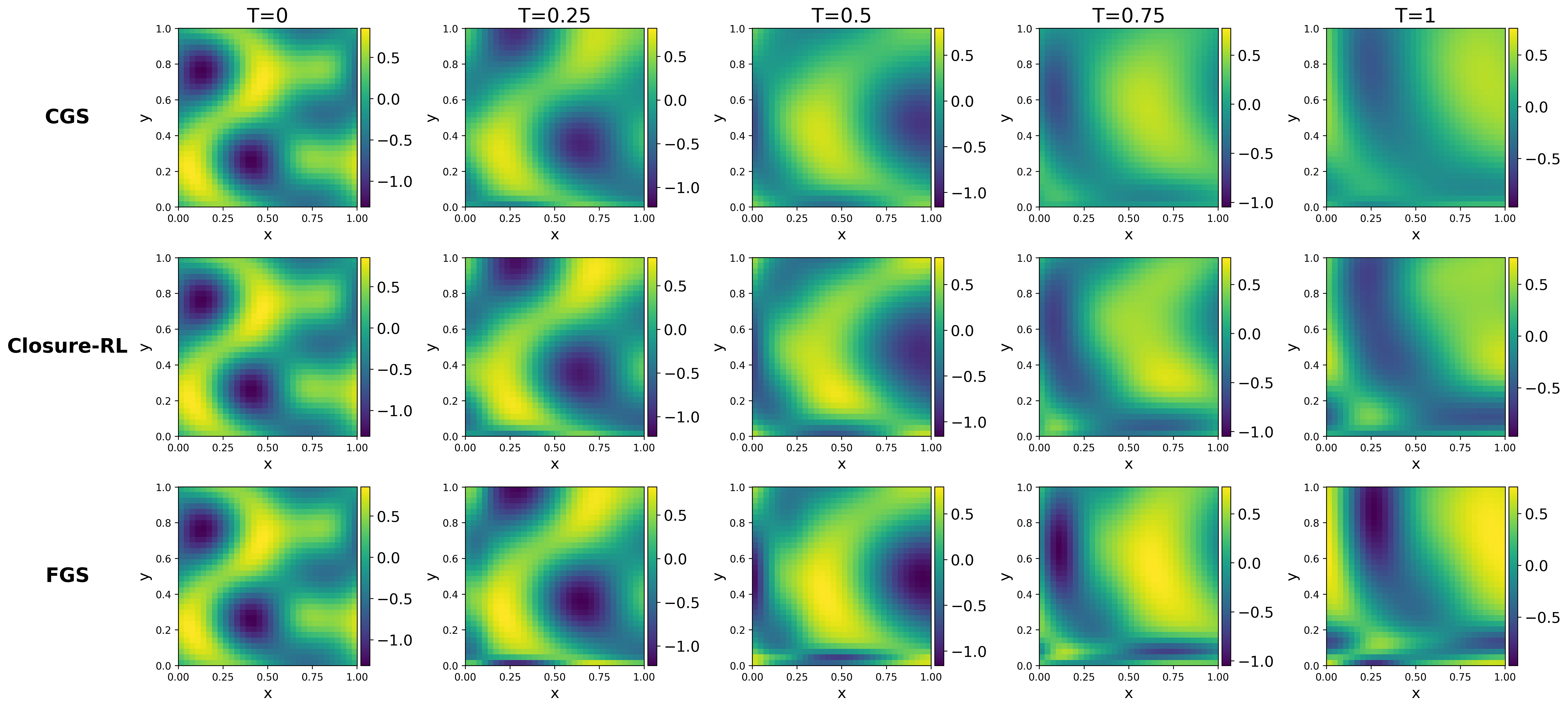}
    \caption{Visualization of the evolution of $\psi$ for the homogeneous 2D advection equation at five different time snapshots. The majority of the error for the CGS solution arises from insufficient magnitudes of the solution. Closure-RL is able to correct this mostly. Comparing the three simulation at $T=0.75$, the dark blue region on the left of the solution domain is not captured perfectly by Closure-RL and even less sufficiently by the CGS when comparing to the FGS.}
    \label{fig:2D_Advection_Unforced}
\end{figure}

\newpage
\section{Comparison to neural operator based approach}
\label{app:fno}
We compared our RL framework to an approach based on a neural operator. In more detail, we used the same synthetic training data for both approaches but instead of learning RL closures, we now learn a Fourier Neural Operator (FNO) \citep{li2020fourier,kovachki2023neural} that directly computes the solution. We note that we did not attempt to learn closures with the FNO directly as this would require a differential CGS solver which can not be guaranteed for many applications.\\
The implementation of the FNO model was provided by the NeuralOperator Python library \citep{kossaifi2024neural}. We choose an architecture with 16 Fourier modes in both the spatial and temporal dimension, with two input channels (for the forcing terms and initial condition), one output channel, 64 hidden channels, and 128 projection channels. The model was trained using a batch size of 50, matching the RL model, with a learning rate of $10^{-4}$, which was found to produce the best performance for this setup (in contrast to the RL model, which used a learning rate of $10^{-5}$).

The training dataset consisted of 14,000 simulation trajectories of the 1D Burgers' equation, each spanning 200 time steps. This approximately matches the number of simulations encountered by the RL model during training: with 1,000 epochs, 2,500 steps per epoch, and an average episode length of 178 steps, the total number of simulated trajectories is approximately 14,000. Regarding validation and test set we employed the same strategy as for the RL training.

The results obtained are displayed in Figure \ref{fig:fno1} for the in-distribution test case (i.e. unseen forcing terms and initial conditions) as well as in Figure \ref{fig:fno2} for the out-of-distribution test case (i.e. homogeneous PDE with unseen initial conditions). We note that while for the first test case the FNO approach results in approximately the same accuracy, the RL closures outperform for the second, extrapolative test cases. We attribute this to the fact that the RL based approach can employ the CGS during the prediction steps, can employ information generated by the CGS and then only uses the closures to correct this CGS. RL closures are thus especially well suited for out-of-distribution prediction tasks.

\begin{figure}[h]
    \centering
    \includegraphics[width=1.0\linewidth]{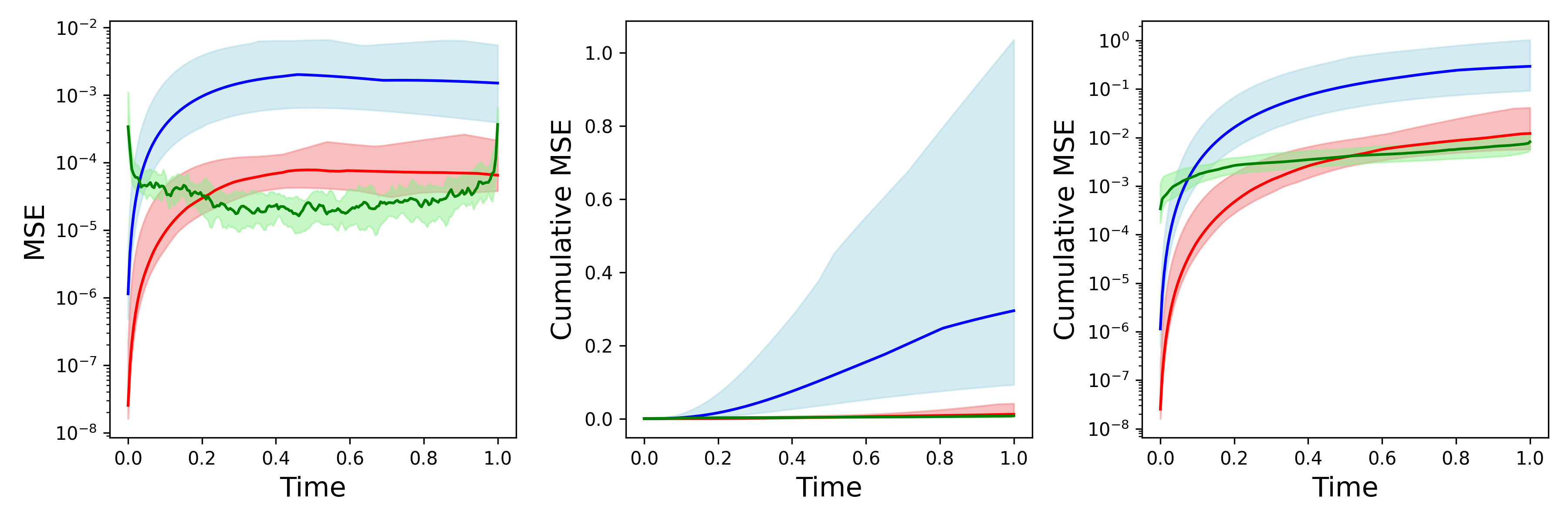}
    \caption{Results for the inhomogeneous 1D Burgers' equation: The mean squared error (MSE) is calculated with respect to the analytical MMS. The plots show median values for the CGS (\textcolor{CGS}{\rule[0.5ex]{1.em}{1.5pt}}), the Closure-RL (\textcolor{RL}{\rule[0.5ex]{1.em}{1.5pt}}) and the FNO solution (\textcolor{green}{\rule[0.5ex]{1.em}{1.5pt}}) across 30 different MMS solutions. The shaded regions around the medians represent the interquartile range (25th to 75th percentile).}
    \label{fig:fno1}
\end{figure}

\begin{figure}[h]
    \centering
    \includegraphics[width=1.0\linewidth]{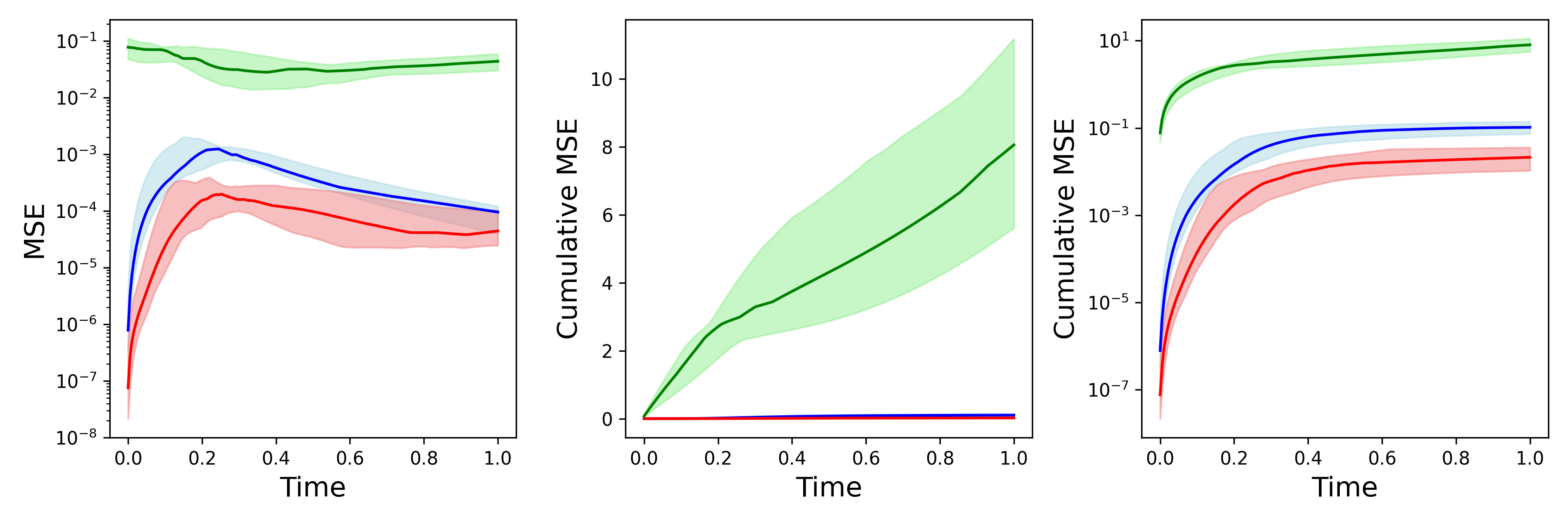}
    \caption{Results for the inhomogeneous 1D Burgers' equation: The mean squared error (MSE) is calculated with respect to a FGS. The plots show median values for the CGS (\textcolor{CGS}{\rule[0.5ex]{1.em}{1.5pt}}), the Closure-RL (\textcolor{RL}{\rule[0.5ex]{1.em}{1.5pt}}) and the FNO solution (\textcolor{green}{\rule[0.5ex]{1.em}{1.5pt}}) across 30 different MMS solutions. The shaded regions around the medians represent the interquartile range (25th to 75th percentile).}
    \label{fig:fno2}
\end{figure}